# ASDA : Analyseur Syntaxique du Dialecte Algérien dans un but d'analyse sémantique

Imène GUELLIL[1]     Faiçal AZOUAOU[2]

[1,2] Laboratoire de méthode de conception de système LMCS
Ecole Supérieure d'Informatique ESI
Alger, Algérie

{i_guellil, f_azouaou}@esi.dz

**Résumé**

*La fouille d'opinions et analyse de sentiments au sein des médias sociaux représente une piste de recherche suscitant un grand intérêt de la communauté scientifique. Néanmoins avant de procéder à cette analyse, nous sommes confrontés à un ensemble de problématiques. La première concerne la richesse des langues et dialectes au sein de ces médias. Afin de répondre à cette problématique, nous proposons dans ce papier une approche de construction et d'implémentation d'un analyseur syntaxique d'un dialecte arabe (le dialecte algérien) nommé ASDA servant à étiqueter les termes d'un corpus donné. Nous récupérons ainsi une table d'étiquetage contenant pour chaque terme son radicale, différents préfixes et suffixes, nous permettant de déterminer les différentes parties grammaticales, une sorte d'étiquetage POS[1]. Cet étiquetage, nous servira par la suite dans le traitement sémantique de ce dialecte, c'est-à-dire la traduction automatique de ce dernier ainsi que l'analyse de sentiment de messages rédigés en ce dialecte*

**Mots Clef**

Dialecte arabe, Dialecte algérien, analyseur syntaxique, analyse de sentiments, lexique du dialecte.

**Abstract**

*Opinion mining and sentiment analysis in social media is a research issue having a great interest in the scientific community. However, before begin this analysis, we are faced with a set of problems. In particular, the problem of the richness of languages and dialects within these media. To address this problem, we propose in this paper an approach of construction and implementation of Syntactic analyzer named ASDA. This tool represents a parser for the Algerian dialect that label the terms of a given corpus. Thus, we construct a labeling table containing for each term its stem, different prefixes and suffixes, allowing us to determine the different grammatical parts a sort of POS tagging. This labeling will serve us later in the semantic processing of the Algerian dialect, like the automatic translation of this dialect or sentiment analysis*

**Keywords**

Dialecte algérien, analyseur syntaxique, analyse de sentiments, lexique du dialecte

## 1 Introduction

L'émergence des médias sociaux a conduit à l'explosion de la quantité de données générées par les utilisateurs. La fouille de ce contenu représente une mine d'or pour plusieurs disciplines et même autre que l'informatique. Néanmoins, confronté à la richesse des opinions, sentiments et émotions présents sur ces médias, une des pistes de recherche de la fouille des médias sociaux suscite un grand intérêt de la part de la communauté de recherche. Cette piste n'est autre que la fouille d'opinion et analyse de sentiments. Cependant, une des caractéristiques majeures du contenu présent au sein des médias sociaux est la richesse des langues et dialectes auxquels font appel les utilisateurs [1]. Récemment, un intérêt considérable a été donné aux dialectes arabes et surtout à la variété de ce dialecte se trouvant au sein des médias sociaux. L'identification et le traitement de dialecte a été même considéré comme le premier composant du prétraitement pour n'importe quel problème NLP (traduction automatique, recherche d'information, sans oublié la fouille d'opinion et analyse de sentiments) [2]. Parmi les dialectes arabes les plus recensé, nous trouvant le dialecte algérien. Ce dernier est caractérisé par l'absence de standard et de ressources [3]. Il diffère de l'arabe classique en termes de représentation linguistique (phonologique et morphologique), et de lexique ainsi que dans la représentation syntaxique. Tous ces aspects rendent les outils NLP qui ont été développé pour le traitement de la langue arabe impuissants devant un tel dialecte [3]. Nous avons cependant constaté que la plupart des travaux effectués se basent sur le dialecte égyptien et tunisien. En ce qui concerne le dialecte algérien très peu de travaux y ont touché [4]. Les travaux ayant traité ce dialecte, se concentrent sur l'aspect linguistique en général en omettant les différentes écritures possibles au sein des médias sociaux. Nous n'avons cependant recensé aucun travail traitant de la syntaxe ou encore du découpage des termes manipulés dans le dialecte algérien. Pour répondre à cette problématique, nous proposons au sein de ce papier un analyseur syntaxique du dialecte algérien. Ce dernier extrait d'un même terme plusieurs parties (correspondant aux verbes, noms, adjectifs, conjonctions, sans oublié les pronoms personnels avec lesquels est conjugué un verbe, pronoms COD, COI, la forme d'un verbe, adjectifs possessifs pour les noms, etc.). Notons seulement que pour construire cet analyseur, nous avons tout d'abord enrichi un lexique de traduction entre le dialecte algérien et le français, à l'aide de différentes extensions phonologiques. A partir de cet analyseur, nous pourrons produire une table d'étiquetage de tous les termes servant par la suite différents traitements sémantiques du dialecte algérien tels que la traduction automatique et l'analyse de sentiments.

---
[1] POS : Part Of Speech

Notre article est organisé comme suit : Nous présentons d'abord les travaux reliés à notre problématique, c'est-à-dire les travaux traitant des dialectes arabes en général et dialecte algérien en particulier en section 2. Nous exposons ensuite notre approche de construction d'analyseur syntaxique au sein de la section 3. Nous implémentons cette dernière à l'aide du langage java au sein de la section 4. Enfin, nous concluons notre article par une synthèse en section 5

## 2 Les travaux étudiés

Plusieurs études ont été menées sur le traitement du langage naturel en arabe. Cependant la plupart des techniques traditionnelles se sont focalisées sur le traitement du MSA (arabe classique) **[2]**. Néanmoins ces dernières années, l'intérêt de traitement des dialectes arabes a augmenté. Cette augmentation peut être attribuée à plusieurs facteurs tels que l'usage étendu des dialectes arabes au sein des médias sociaux **[4]**. Pour bien présenter les choses, nous allons diviser cette section en deux parties : la première présentant les travaux sur le traitement des dialectes arabes en général, la deuxième présentant les travaux menés sur le dialecte algérien en particulier.

### 2.1 Traitement des dialectes arabes en général

En se basant sur l'étude de l'état de l'art **[4]**, nous classons les travaux menés sur le ANLP (traitement de langage naturel du dialecte arabe), en quatre catégories principales : 1) Analyse basique du langage, 2) Construction des ressources pour les dialectes arabe, 3) Identification des différents dialectes et 4) Analyse sémantique.

L'analyse basique du langage consiste à étiqueter les différentes partie d'un corpus à l'aide d'une analyse morphologique ou encore l'analyse syntaxique et orthographique. Ces travaux veillent cependant au respect des différentes règles syntaxiques. Plusieurs travaux ont été menés au sein de cette catégorie, parmi ces derniers nous citons les travaux de HABECH et al. dans [5], qui présentent un analyseur et générateur morphologique, nommé MAGED. Cet outil peut analyser le dialecte levantine (le dialecte de la Syrie, le Liban, l'Israël, la Palestine ainsi que la Jordanie) et convertir par la suite le MSA (l'arabe classique) au levantine. Les règles orthographiques de cet outil ont été détaillées par la suite dans l'article de HABASH et al. dans **[6]**. En ce qui concerne l'analyse syntaxique, SADAT et al. dans **[7]** présentent un ensemble de règles grammaticales fournissant des ressources de traduction du dialecte tunisien au MSA ainsi qu'à d'autres langues définis. Pour l'analyse orthographique, les travaux s'orientent vers la proposition de conventions orthographiques servant à construire des modèles pour le traitement des dialectes arabes tel que l'égyptien **[8].**

Les travaux s'orientant sur la construction de ressources pour les dialectes arabes concernent : la construction de lexiques et de corpus. La construction de lexique se fait en premier lieu manuellement, leurs enrichissement se fait par la suite à l'aide de dictionnaire de langues tels que l'arabe, l'anglais, etc. Ceci à pu être fait pour la construction d'un lexique irakien par GRAFF et al. dans**[9],** ainsi que pour le lexique tunisien par BOUJELBANE et al. dans **[10].** La construction de corpus quant à elle inclue l'identification des règles de bases (tels que les différentes écritures de voyelles). Cette construction fait appel à une autre catégorie de travaux qui n'est autre que l'identification de dialectes **[11]**. Ces corpus peuvent également être construits à l'aide de corpus parallèle, comme cela a été fait pour le dialecte algérien par SMAÏLI et al. dans **[10]**.

L'identification des différents dialectes a pu être effectuée dans deux axes de recherches (textuelle et vocale). Dans le cadre de l'identification textuelle, ELFARDY et al. dans **[13]** propose une approche supervisée basée sur le niveau phrase afin de différencier entre l'arabe classique et le dialecte égyptien. L'équipe de SADAT et al. considèrent que le problème d'identification des différents dialectes arabe peut être traité à l'aide d'un modèle du langage Markovien N-gram **[2]**. Ce modèle est utilisé pour calculer la probabilité qu'un texte en entré soit dérivé d'un langage donné. Pour la classification des différents dialectes arabes, ces auteurs font appel à un classificateur bayésien naîf. En ce qu'est de la reconnaissance vocale du dialecte, la plupart des travaux menés s'orientent vers la reconnaissance vocale en premier lieu pour ensuite identifier le dialecte **[14]**.

Pour la dernière catégorie de travaux gérant l'analyse sémantique des dialectes arabes, la plupart des auteurs se consacrent au traitement de la traduction. De ce fait, nous pouvons citer les travaux de MOHAMED et al. dans **[15]** où les auteurs présentent un traducteur de l'arabe classique vers le dialecte égyptien. Dans l'article de JEHL et al. dans **[16]**, les auteurs ont collecté un ensemble de pairs de phrases bilingues pour entrainer un système de traduction statistique. Ceci afin de traduire des messages de *microblogs* dans les différents dialectes des pays du golf, levantine, égyptien, etc. D'autres tâches sémantiques consistent à utiliser le traitement du dialecte arabe pour l'analyse de sentiments et de subjectivité. Par exemple, dans **[17],** les auteurs se focalisent sur le traitement de dialectes arabes, plus précisément sur la manière d'extraire les caractéristiques ayant le plus grand impact sur l'analyse de sentiments. Dans **[18]**, les auteurs présentent un classificateur de sentiment des expressions courantes du dialecte arabe. Ces derniers se basent sur les commentaires extraits de *Facebook*.

Les auteurs de cet état de l'art **[4]**, terminent leur analyse en présentant un tableau synthétisant les travaux effectués dans chacune des quatre catégories tout en distinguant entre les différents dialectes traités. A partir de ce dernier, nous pouvons constater que la plupart des travaux menés concernent les dialectes : égyptien, levantine, tunisien et irakien. Néanmoins, nous n'avons pu recenser sur ce tableau que trois travaux concernant le dialecte algérien.

## 2.2 Traitement du dialecte Algérien

En ce qui concerne les travaux effectués sur le dialecte algérien, nous pouvons en premier lieu citer le travail de SAADANE al dans **[3]**, qui peut être classé au sein de la catégorie traitant de l'analyse basique des dialectes arabe et plus précisément celle se concentrant sur la présentation de règle orthographiques. Les auteurs de cet article se basent sur le modèle orthographique CODA [2] (Conventions orthographiques pour les dialectes arabes) proposé par HABASH et al. dans **[8],** pour le dialecte égyptien. Ces auteurs débutent leur travail en comparant le dialecte algérien au dialecte tunisien et égyptien. Ils présentent par la suite les différentes variations phonologiques entre le dialecte algérien, égyptien et tunisien, tel que le *q, g, ʔ,* ou *k* pour la lettre arabe (ق), le dj pour la lettre (ج) ou encore *γ pour*(غ) ou θ pour (ث). Ces auteurs proposent également l'ajout des deux lettres (yn) comme suffixes aux noms pour former leurs pluriels ainsi que la présence du caractère n en début des verbes qui prévoit que le verbe est conjugué à la première personne du singulier. Ils finissent par les différentes variations lexicales. Ils présentent ensuite le CODA algérien tout en présentant les différentes extensions effectuées au CODA égyptien.

Pour les travaux sur la construction de ressources, nous recensons deux travaux de MEFTOUH et al. dans **[19]** et SMAÏLI et al dans **[12]**. Le travail de MEFTOUH et al. est certainement le premier des travaux ayant traité le dialecte algérien. Cet article présente une analyse linguistique d'une catégorie de dialecte algérien qui n'est autre que le dialecte d'Annaba (Est de l'Algérie). Il expose également la méthodologie suivie par les auteurs afin de construire un corpus parallèle entre l'arabe classique (MSA) et le dialecte arabe. Ce corpus servira par la suite à la traduction automatique des textes écrits en dialecte arabe vers le MSA. Pour cela, ces auteurs présentent les spécifications du dialecte d'Annaba. Ils présentent également les généralités reliées à ce dialecte tel que la traduction des pronoms personnels, l'ajout des suffixes, la formulation de la phrase interrogative et négative en dialecte d'Annaba, etc. Le deuxième travail **[12]**, représente la suite du premier. Le but de ce dernier est de procéder à une traduction entre le dialecte algérois et le MSA. Ce dernier se consacre sur deux types de dialectes : l'Algérois et le dialecte d'Annaba. Pour ce faire, les auteurs construisent deux dictionnaires (le premier MSA-Algérois et le deuxième MSA-Annaba) et ce à partir des sites de diffusion de film et d'émissions Algériennes. Ils construisent par la suite un dictionnaire du dialecte se basant sur les différences existantes entre les dialectes Algérien et le MSA (tels que les préfixes au début des verbes, le pluriel des noms féminin, etc). Ces auteurs finissent leur travail en exposant le premier moteur de traduction entre le MSA et dialecte Algérien. Cependant, ce dernier ne donne pas de très bons résultats (surtout pour l'Algérois). Les auteurs soulèvent des difficultés de traduction du dialecte algérois (plus que le dialecte d'Annaba). Ces derniers expliquent cela par le fait que le dialecte d'Annaba est beaucoup plus proche de l'arabe classique MSA que le dialecte algérois.

---

[2] Convention Orthography for Dialectal Arabic

Après analyse de ces travaux nous constatons les problématiques suivantes :
- Le manque de papiers traitant du dialecte algérien ainsi que la non couverture de plusieurs catégories de recherche pour ce dialecte (celles concernant l'analyse syntaxique, construction de lexique ainsi que les différentes analyses sémantiques).
- Dans l'analyse orthographique des dialectes algériens ([3] et [19]), les auteurs présentent certaines conventions entre les lettre arabes et celle utilisées en dialectes tels que *γ pour* (غ) ou θ pour (ث). Néanmoins ces conventions sont valables d'un point de vue linguistique seulement. Dans le cas d'exploitation, de ces connaissances dans le cadre des médias sociaux, elles ne donneraient pas un bon résultat, puisque aucun des utilisateurs de ces médias ne fait appel au caractère *γ pour* (غ) et θ pour (ث).
- Aucun des travaux effectués sur le dialecte algérien ne prévoit l'exploitation de ces résultats dans le cadre d'analyse de sentiments dans les médias sociaux. De ce fait les concepts présentés, les verbes analysés sont très générales et non dirigés vers ce type d'analyse particulière.

Nous présentons dans la section suivante notre approche visant à résoudre les problèmes identifiés ci-dessus.

## 3 Contribution

Pour répondre aux problématiques citées précédemment, nous proposons une approche d'analyse et de traitement du dialecte algérien. Cette approche analyse le dialecte algérien au sein des médias sociaux dans le but de construire un analyseur syntaxique des termes d'un corpus donné.

Nous visons plusieurs objectifs par cette approche. D'abord nous comptons analyser les termes du dialecte algérien utilisé dans les médias sociaux. Pour ce faire, nous nous appuyons sur un corpus de données extrait précédemment. Nous extrayons par la suite le modèle de termes de ce corpus, qui pourrait également être utilisé dans de futurs travaux, dans le cadre d'enrichissement de lexique du dialecte algérien. L'analyse effectuée servira plusieurs traitements dont le prétraitement des termes, l'extraction et enrichissement de lexique ainsi qu'à la construction d'un analyseur syntaxique servant à étiqueter les termes d'un corpus donné. L'analyseur syntaxique permettra la construction d'une table d'étiquetage des termes d'un corpus. Notons que la table d'étiquetage de tous les termes du corpus nous servira par la suite dans de futurs travaux où nous nous intéresserons à l'analyse sémantique du dialecte algérien dans les médias sociaux. Par aspect sémantique, nous désignons la traduction automatique ainsi que l'analyse de sentiments. Afin d'éclaircir la situation, nous divisons notre contribution en deux parties : 1) analyse du dialecte algérien, analysant les termes les plus utilisés au sein des médias sociaux et 2) traitement de ce dialecte en présentant un analyseur syntaxique des différents termes. Nous détaillons ces deux parties dans ce qui suit :

## 3.1 Analyse du dialecte Algérien au sein des médias sociaux

Pour pouvoir effectuer cette analyse nous nous basons sur un corpus de données que nous avons extrait dans un précédent travail, de la page de *Facebook*[3] (reconnu être le média social le plus populaire) [1]. Notre premier constat est que les algériens au sein de cette page (ainsi que dans d'autres pages), font appel au dialecte algérien ainsi qu'au français pour communiquer (rarement à l'anglais). Ce qui fait appel à une autre problématique reconnu par le « *code Switching* » (c'est-à-dire faire appel à plusieurs langues et dialecte au sein du même discours. Nous pouvons donc trouver des commentaires de la forme : « *inchlah nafozo ma3kom* » (en dialecte, signifiant, si dieu le veux, nous gagnerons avec vous), « Dommage qu'en Algérie il ne marche que par défaut en plus du fait qu'il ne fonctionne pas en mode hors connexion » (en français) ou encore « mais nchallah ghadi nab9a ntélécharger ghi manah » (un mélange entre le français et le dialecte algérien, signifiant mais si dieu le veux nous n'allons plus télécharger que ce de cet opérateur). Afin d'étudier l'ensemble des mots utilisés dans ce corpus, nous avons en premier lieu construit un modèle de termes, qui contient pour chaque terme sa fréquence d'utilisation dans le corpus (dans le but de trier les termes par rapport à leur fréquence d'apparition). Nous avons ensuite filtré les termes pour ne garder que ceux appartenant au dialecte algérien (avec leurs fréquences d'utilisation bien sûre). Nous pouvons illustrer les étapes de cette partie au sein de la "Fig. 1".

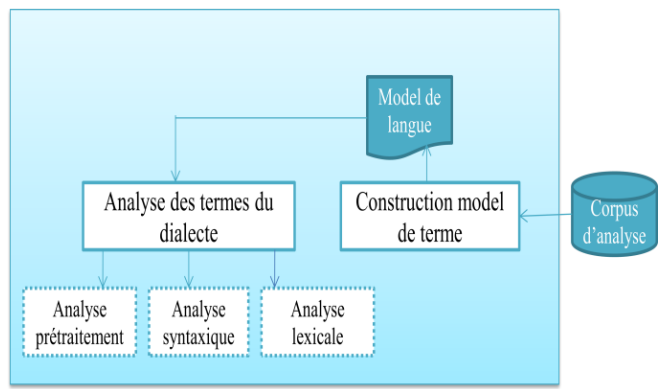

Fig .1. Analyse du dialecte algérien dans les médias sociaux

En se penchons sur les termes appartenant au dialecte, nous pouvons extraire un ensemble de constats:

- Le premier constat concerne le besoin de prétraitement de certains termes. Par exemple nous remarquons que les termes bezzzzzzaf, Sahiiiit, sahbiii contiennent la répétition d'une certaine lettre (z dans le premier terme, i dans le deuxième et troisième). Des termes sous cette forme ne peuvent donc pas être reconnus par le lexique avant que cette exagération ne soit traitée.

- Le deuxième constat concerne l'analyse phonologique de certains sons utilisés dans les médias sociaux. Cette analyse servira à enrichir notre lexique de dialecte-français. Les sons détectés sont les suivants : Pour le son (غ), les utilisateurs font appel à la combinaison des deux lettres (gh), par exemple : *ghaya, ghalta, ghiba*, etc. En ce qui concerne le son (ق), la plupart des travaux étudiés font référence aux différentes lettres k et q. Pour notre part nous constatons l'apparition du chiffre (9) pour remplacer ce son, par exemple: *wa9tach, na9sin, t9oul*. Cet appel aux chiffres remplaçant les lettres se fait beaucoup ressentir au sein des médias sociaux, par exemple pour la lettre (ع), les utilisateurs font appel au chiffre (3) (*3yana, ma3kom, ma3andek*), ou encore pour la lettre (ح), qui est représentée par le chiffre 7 (*Ch7al, sa7, 7ala*). Noter seulement que la lettre (ع), peut également être représenté à l'aide des deux lettres (aa) et (ح) à l'aide de h (Néanmoins h peut également représenter la lettre (ه) comme dans *hayla*). Pour le son (ش), il peut être représenté par ch (*machi, chokran, Manich*), che (*makache, matgoloche, matahbasche*) ou encore sh (*kash, Wesh, bash*). Le (و) peut également être écrit de différentes manières, à l'aide du w (*wa9tach*), des deux lettre oua (*houa*), oui (*touil*). Le son (ي), peut être obtenu à l'aide de la lettre y (*khoya, kayna, yi*). Le son (خ), peut être obtenu grâce aux deux lettres kh(*khoya, khir*). Le son (ج), peut être représenté soit à l'aide de la lettre j (*haja, nji*) ou encore la combinaison des deux lettre dj (*hadja, ydji*). La dernière remarque que nous pouvons faire concerne les voyelles en arabe ou en dialecte (reconnu par *tachkil*). Nous ne parlerons, néanmoins que d'*el fatha* (qui s'effectue à l'aide d'un petit trait sur la lettre comme dans بَ). Nous avons cependant constaté que cette *fatha* pouvait s'écrire en dialecte, à l'aide de « a » (*rabi*) ou « e » (*rebi*). De même pour la *damma* (tel que dans بُ). Cette dernière est représentée soit à l'aide de « o » (*khoya*) ou « ou » (*khouya*).

- Le troisième et dernier constat concerne l'analyse syntaxique des différents termes. Contrairement aux autres langues, l'analyse syntaxique, peut concerner un seule terme. Par exemple, la phrase « Je t'ajoute… » en français, est reconnu au sein de notre dialecte par « *nzidlek* » où le n représente je, *zid* représente le verbe « ajouter » et *lek* représente le pronom COI « te ». Nous pouvons citer un autre exemple concernant les noms, la phrase « mon dieu » sera représenter par « *rabi* » en dialecte. Ce terme peut donc être divisé en deux parties : nom qui est « *rab* » et l'adjectif possessif mon qui est « *i* ». De même pour les adjectifs où le « a » à la fin désigne le féminin. Par exemple pour « *sghira* », vu que « *sghir* » désigne petit, donc en ajoutant le « a » à la fin, ça désignera « *petite* ».

---
[3] https://www.facebook.com

Les différents constats auxquels nous avons pu aboutir au sein de cette étape, nous servirons pour concevoir les différentes techniques et algorithmes des prochaines étapes.

## 3.2 Traitement du dialecte Algérien au sein des médias sociaux

Cette partie reçoit en entré un corpus en dialecte algérien devant être traité dans le but de concevoir une table d'étiquetage où chaque terme de ce lexique sera étiqueter. Le but principal de cet étiquetage est de pouvoir extraire différentes parties grammaticales encapsulées dans le même mot. Cette table pourra donc nous servir dans de futurs travaux afin de procéder à plusieurs analyses sémantiques telles que la traduction ou l'analyse de sentiment. Néanmoins pour pouvoir aboutir à cet analyseur syntaxique, le terme doit être préalablement prétraité et un lexique de dialecte doit être construit et enrichit à l'aide de règles phonologiques. Nous illustrons les étapes de cette partie au sein de la "Fig. 2". Afin d'éclaircir les choses, nous présentons au sein de cette partie trois aspects : 1) Le prétraitement de termes, 2) la construction et enrichissement d'un lexique dialecte-français et 3) la construction d'un analyseur syntaxique.

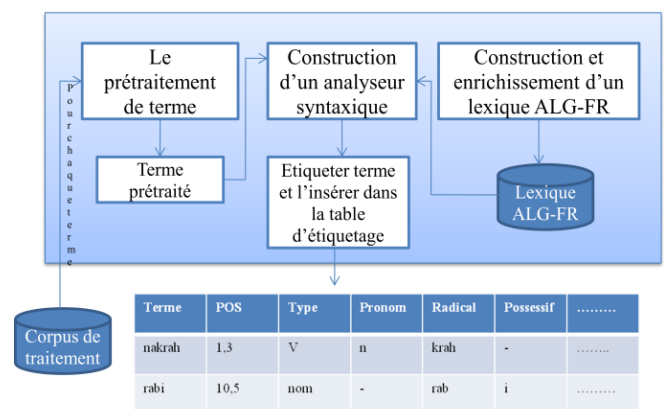

Fig .2. Traitement du dialecte algérien dans les médias sociaux

*1) Le prétraitement des termes :* D'après les résultats d'analyse effectués précédemment, rappelons qu'un terme peut contenir des exagérations (bezzzzzzzzzzzaf). Avant de procéder à l'étiquetage des termes, nous devons tout d'abord vérifier que le terme n'est pas dans cette catégorie. Notons que la recherche de répétitions de lettre peut se faire à l'aide des expressions régulière. Pour les termes contenant des exagération, nous devons tous d'abord supprimer les lettres répétés. Nous devons également garder trace des mots contenant des exagérations car ceci ne pourra qu'accentuer le sentiment.

*2) La construction et enrichissement d'un lexique dialecte-français:* S'appuyant sur les caractéristiques du dialecte algérien présenté au sein de l'analyse lexicale, nous avons pu extraire du net[4] un lexique contenant la traduction entre les termes du dialecte algérien et le français. Ce dernier contient 228 verbes, 73 adjectifs, 297 noms, ainsi que d'autre nom et expressions particulières (reliés au domaine médical et autre). Dans le cadre de ce travail, nous nous concentrons sur les verbes, adjectifs et noms. Nous laissons la partie domaine particulier et expressions particulière, à traiter dans d'autre travaux. En se penchant sur ce lexique, nous avons constaté qu'il a été effectué dans un but linguistique, la représentation de lettre en français est donc loin de celle utilisée au sein des médias sociaux. Par exemple le « ح » a été représenté par « ḥ », le « غ » par « ġ », le « ش » par « š », etc. Néanmoins, d'après tous les exemples que nous venons de citer précédemment, nous pouvons constater qu'au sein des médias sociaux, aucun utilisateur ne fait appel à ces lettres pour s'exprimer. La première étape de l'amélioration de ce lexique est donc de changer ces lettres par les lettres les plus utilisés au sein des médias sociaux.

Vu que le but primaire de notre travail est de pouvoir traiter le dialecte algérien dans un contexte d'analyse de sentiment. Nous devons donc intégrer des termes représentant le sentiment de la population à notre lexique. Dans un premier temps, nous comptons uniquement extraire les verbes, adjectifs et noms reconnus avoir une polarité intense au sein de SentiWordNet [5] (tels que excellent, lamentable, love, disrespect, hate,…), sans oublié leurs traduction vers le dialecte algérien. Notons seulement que SentiWordNet est une ressource très utilisés dans les travaux de recherche traitant de l'analyse de sentiment.

D'après l'analyse du modèle de termes construit précédemment, nous pouvons constater l'apparition de plusieurs termes dont la fréquence est élevée. Néanmoins ces derniers ne sont pas inclus au sein de notre lexique. Parmi ces derniers nous pouvons citer : « *fi* » qui signifie « dans », « *ana* » qui signifie « moi », « *nchallah* » qui signifie « si dieu le veut », etc. nous définissons donc une partie de notre lexique reconnue par (particules : conjonctions et pronoms) où nous définissons la traduction de ces termes.

Néanmoins, comme nous l'avons cité précédemment, un même terme peut s'écrire différemment selon l'utilisateur. Après analyse des différents termes ayant le même sens mais qui diffèrent au sein d'une lettre ou deux, nous pouvons extraire des règles de convention entre les lettres nous permettant d'enrichir le lexique obtenu. Ces conventions ne sont autre que l'enrichissement des mots contenant le son « q » par « k et 9 », le son « k » par « q et 9 », le son « h » par 7 ou encore le son « ch » par « che et sh », ect.

---

[4]http://www.flipsnack.com/95C5C758B7A/dictionnaire-arabe-algerien.html

[5]http://sentiwordnet.isti.cnr.it/

*3) La construction d'un analyseur syntaxique :* Après une analyse détaillée de la structure des éléments : verbes, noms et adjectif, nous avons pu dégager une syntaxe générale du dialecte algérien les concernant.

Pour les verbes, nous nous intéressons au mode impératif et indicatif. Notons seulement que pour le mode indicatif, nous nous concentrons, que sur le temps présent (nous comptons étendre cet analyseur au autres temps dans de futurs travaux). Nous nous penchons également sur la négation (ne…pas), les pronoms COD (me, le, la, les…) ainsi que les pronoms COI (me, lui, leur,…). Vu que notre recherche sera orientée par la suite vers l'analyse de sentiments dans les médias sociaux, prenant par exemple deux verbes exprimant un sentiment et une émotion tels que : aimer et pleurer. La TABLE.I ci-dessous, présente les différents changements que présentent ces deux verbes pour l'impératif et le présent ainsi que pour la négation et les différents pronoms

TABLE. II. Traitement syntaxique des verbes en dialecte algérien

| Verbe/ préfixes et suffixes | Le mode impératif | Présent à l'aide des pronoms n/t/y/i | Négation Ma…ch | Pronoms COD ni/ek,k/ou/h,ha/na/ kom/hom | Pronoms COI Li/ lek/lou,lha/ena, elna,lna/ elkom/lkom/ elhom/lhom |
|---|---|---|---|---|---|
| Aimer | hab habi habou | *n*hab *t*hab/ *t*hab*i* *y*hab/ *i*hab/ *t*hab *n*hab*ou* *t*hab*ou* *y*hab*ou* | *ma*n*hab*ch* *mat*hab*ch*/ *mat*hab*ich* *may*bab*ch*/ *mat*hab*ch* *ma*n*hab*ouch* *mat*hab*ouch* *may*hab*ouch* | yhab*ni* yhab*ek* yhab*ou*/ yhab*ha* yhab*na* yhab*kom* yhab*hom* | yhab*li* yhab*lek* yhab*lou* yhab*elha* yhab*ena/* yhab*elna* yhab*elkom* yhab*elhom* |
| Pleurer | ebki ebki*ou* ebki*w* | *n*ebki *t*ebki *y*ebki/ *t*ebki *n*ebk*ou* / *n*ebki*w* *t*ebk*ou*/ *t*ebki*w* *y*ebk*ou* / *y*ebki*w* | *ma*n*ebki*ch* *mat*ebki*ch*/ *may*ebki*ch*/ *t*ebki*ch* *n*ebk*ouch*/ *n*ebki*wch*/ *mat*ebk*ouch*/ *mat*ebki*wch* *may*ebk*ouch*/ *may*ebki*wch* | yebki*ni* yebki*k* yebki*h*/ yebki*ha* yebki*na* yebki*kom* yebki*hom* | yebki*li* yebki*lek* yebki*lou* yebki*lha* yebki*lna* yebki*lkom* yebki*lhom* |

Pour les noms, nous nous intéressons à la combinaison de ces derniers avec les adjectifs possessifs. Néanmoins pour pouvoir généraliser la terminaison, nous identifions plusieurs cas en rapport au genre et au nombre du nom ainsi que le genre et le nombre de l'adjectif possessif. Par exemple pour le nom « *ktab* » (qui veut dire livre), l'ajout du "i" à la fin dans *ktabi* signifie mon livre, l'ajout du kom dans ktabkom signifie votre livre, etc.

Concernant les adjectifs, nous n'avons pu faire cette généralisation que pour le féminin où l'adjectif prend « a » à la fin pour passer du masculin au féminin. Nous pouvons prendre l'exemple de l'adjectif « *sghir* : petit » qui deviendra « *sghira* » au féminin ou encore « *meskine* : pauvre » qui deviens « *meskina* ». Nous présentons à la"Fig. 3"., notre analyseur syntaxique sous forme d'automates à états finit. Cet automate contient 15 états dont 10 qui peuvent être finaux. Pour ne pas encombrer cet automate, nous préférons représenter les transitions par leurs dénominations générales. Par exemple par première partie de la négation, nous référons aux « *ma* », par pronoms personnels, nous distinguons : n/t/y/i/ne/te/ye/na/ta/ya. Notons seulement que les auxiliaires être et avoir ne suivent pas le même model que les autres verbes et ceci est parfaitement représenté sur notre automate.

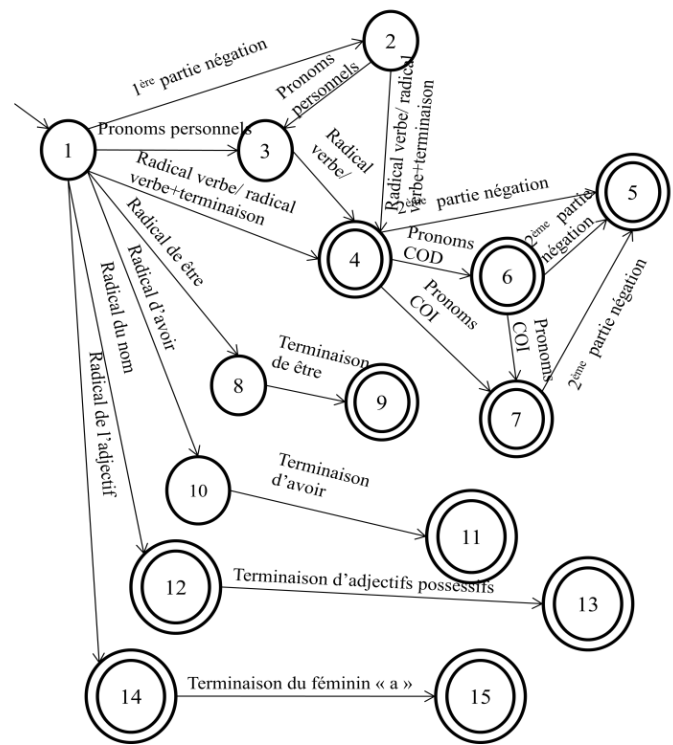

Fig. 3. Automate représentant notre analyseur syntaxique

Afin d'étiqueter un terme du corpus, il nous suffit de récupérer le chemin de l'automate qu'il est entrain de suivre. Supposons qu'on ait le terme « *mandirhach* ». Nous devons donc commencer par l'état initial (1). Nous passerons ensuite par la transition qui reconnait « ma » pour aboutir à l'état (2). Nous passerons par la transition qui reconnait « n » pour aboutir à l'état (3). Nous chercherons ensuite le radical « dir » dans le lexique, comme ce dernier existe, nous passons à l'état (4). Nous passons ensuite par la transition « ha » pour aboutir à l'état (6) (car ha est un pronom COD). Nous finissons par la transition qui clôture la négation, donc « ch » qui nous fait arriver à l'état (5) (qui est un état final). Ce terme est donc reconnu par l'automate. A l'aide de cet automate nous pouvons également garder trace des étapes par lesquelles nous sommes passés et donc étiqueter le terme (qui a été reconnu).

A travers ces différents passage, l'étiquetage de ce terme sera de cette forme *mandirhach* (type : verbe, radical :dir, pronom personnel : n, pronom_COD :ha, pronom_COI :-, négation : true). L'étiquetage des termes combiné à des règles de traduction nous permettra dans de futurs travaux à procéder à la traduction d'un corpus du dialecte algérien vers une autre langue. Cet étiquetage nous servira également au sein de l'analyse de sentiments des utilisateurs des médias sociaux vis-à-vis d'un sujet donnée.

## 4 Implémentation

Notre approche a été implémentée au sein de l'environnement Windows 7, à l'aide du langage de programmation java (JRE 8 SE), sous l'IDE Eclipse. En ce qui concerne le stockage des lexiques et corpus étudiés, nous utilisons le tableur Excel 2007. Nous n'avons pas besoin d'utiliser une base de données puisque le nombre de nos enregistrements est pour l'instant limité. Néanmoins, nous comptons par la suite utiliser une base de données Nosql pour le stockage de nos corpus extraits des médias sociaux. Afin d'implémenter notre approche, nous avons procédé comme suit :

Nous enrichissons, en premier lieu, notre lexique de traduction entre le dialecte algérien et le français. Pour cela nous faisons appel aux différents termes de fortes intensités extraits de « SentiWordNet », tels que : aimer, détester, splendid, etc. En premier temps, nous extrayons les termes de « SentiWordNet » ayant une valence (qu'elle soit positive ou négative) égale à 1 (100%). C'est-à-dire les termes étant strictement positif ou strictement négatif. Nous procédons ensuite à la traduction manuelle de ces termes vers le Français ainsi que vers le dialecte Algérien .Nous ajoutons une partie particules (conjonction et pronoms) à notre lexique de base, tels que : dans, de, moi, etc. Nous finissons par ajouter les différentes extensions phonologiques permettant à chaque mot d'avoir différentes manière d'écriture par exemple : *oq3od* qui est enrichie en *oq3oud, ok3od, ok3oud, ouq3od, ouq3oud, ouk3od, ouk3oud, o93od, o93oud*. Pour aboutir à tous ces termes, nous avons simplement procéder aux enrichissements phonologiques présentés ci-dessus. Pour ce cas, nous étendons l'apparition de « o » par « ou » et « q » par « k » et « 9 ». Notons que pour un seul terme nous obtenons 9 autres termes.

Pour le corpus de traitement, nous nous sommes basés sur un ensemble de messages extrait sur la page *Facebook* de l'opérateur de téléphonie Algérien « *Mobilis* »[6]. Nous avons par la suite extrait manuellement les messages rédigés en dialecte algérien (notez que nous comptons par la suite automatiser cette tâche à l'aide d'un travail sur la détection du dialecte algérien au sein des messages). Nous présentons au sein de la "Fig. 4", cinq de ces messages.

Pour le traitement de ces messages, nous devons d'abord, découper chaque message en mots. Chaque mot sera ensuite analysé pour voir s'il contient une exagération (répétition d'un caractère pour accentuer un sentiment) de la forme : *maroooooooooooooook*. Une fois le terme détecté, cette exagération sera supprimée à l'aide des expressions

[6] http://www.mobilis.dz/

régulières. Le terme obtenu peut par la suite être analysé syntaxiquement.

Pour procéder à l'analyse syntaxique d'un terme, nous avons tout d'abords implémenté l'automate en construisant en premier lieu sa table de transition [20,21] (considéré également comme la matrice d'adjacence dans le cas des graphes). Cette dernière a ensuite été implémentée à l'aide d'un algorithme vérifiant l'appartenance d'un terme à notre dialecte et en étiquetant chaque partie des termes acceptés. Le but de ce travail est de pouvoir récupérer une table d'étiquetage de toutes les parties syntaxiques et ce de tous les termes reconnus par notre automate. Par exemple, nous pouvons constater que le terme *tro7*, se compose d'un radical « *roh* » qui fait partie des verbes (ce dernier signifie aller en français). Ce terme est également composé d'un préfixe « t » (indiquant le pronom personnel associé à ce verbe). Il ne contient cependant aucune exagération, aucun préfixe ou suffixe de négation et aucun pronom COD, ni même COI.

Nous présentons au sein de la "Fig. 4", ci-dessous, les mots reconnus à partir des messages présentés, ainsi que leurs différents découpages syntaxiques. Notez seulement que pour chaque mot, nous devons également garder sa position (c'est-à-dire le numéro de message auquel il appartient ainsi que sa position au sein de ce message). Pour des besoins de clarté d'affichage, nous présentons au sein de cette maquette un regroupement des différentes tables d'étiquetage selon la partie grammaticale (c'est-à-dire que nous présentons une table pour les verbes, une pour les noms, une autre pour les adjectifs et la dernière pour les conjonction et pronoms).

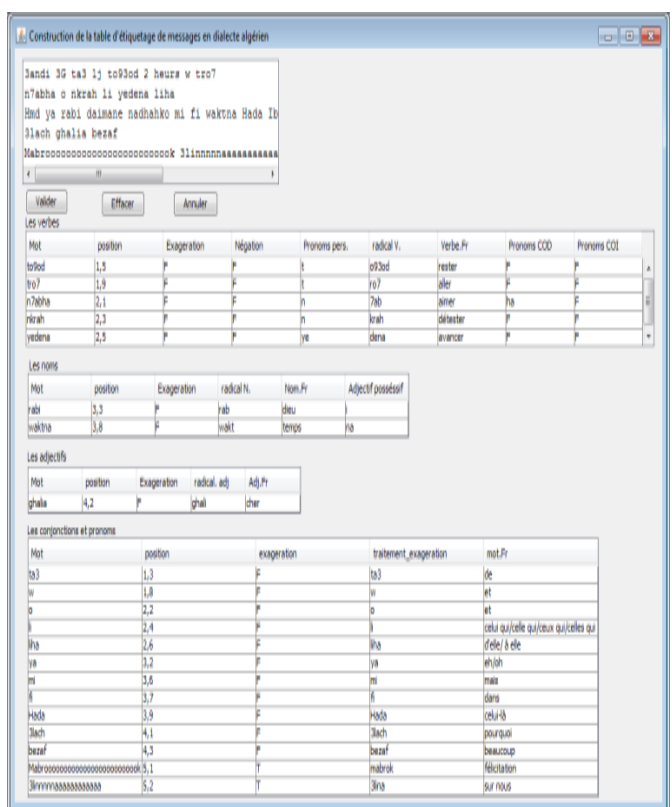

Fig. 4 Interface de l'application

Cette table d'étiquetage est la première résultante d'un grand travail de recherche se consacrant à la fouille d'opinion et analyse de sentiments des commentaires au sein des média sociaux en particulier et sur l'intégralité du web en général. Pour cela, nous nous basons sur la première des problématiques citée au sein de notre précédent travail dans [1], c'est-à-dire concernant le traitement des différentes langues et dialectes (plus précisément, le dialecte Algérien).

## 5 Conclusion

L'objectif de ce travail est de présenter, ASDA, un analyseur syntaxique du dialecte algérien. Ce dernier permettra par la suite, la traduction automatique et l'analyse de sentiments de messages rédigés en ce dialecte. Pour ce faire, nous nous sommes d'abord concentrés sur l'analyse des dialectes arabes et plus précisément le dialecte algérien. Contrairement aux autres travaux [3,19,12], nous avons considéré un environnement particulier pour l'utilisation de ce dialecte qui n'est autre que les médias sociaux. Nous avons analysé le style d'écriture au sein de ces médias, puis proposé un certain nombre de formalismes nous aidant à extraire et enrichir un lexique de traduction entre le dialecte algérien et le français. Une fois ce lexique construit nous nous sommes concentrés sur l'analyse syntaxique des différents termes d'un corpus donné au sein des médias sociaux. Pour ce faire nous avons proposé un analyseur syntaxique sous forme d'automate aidant au découpage et étiquetage des termes. Enfin, nous avons présenté la maquette d'implémentation de notre approche, prenant en entrés un ensemble de messages en dialecte algérien et dégageons une table d'étiquetage des différents termes de ces messages. Cette table servira par la suite à la traduction automatique et analyse de sentiments des messages en dialecte algérien. Cette approche présente cependant un certain nombre de failles qui peuvent être amélioré à l'aide des points suivants :

- Intégrer à cette approche une partie de détection automatique du dialecte algérien, ce qui nous permettra de filtrer les messages écrit en dialecte des autres messages rédigés en français ou en anglais.
- Cette approche ne traite que les verbes en mode impératif et indicatif présent, ça serait intéressent d'intégrer les autres temps, tels que le passé et le futur, etc.
- Cette approche procède à une analyse terme par terme, or que le dialecte algérien est riche en terme d'expression. Il serait donc intéressent d'intégrer une analyse niveau phrase et non niveau terme.

## Bibliographie

[1] Guellil, Imene;Boukhalfa, Kamel, "Social big data mining: A survey focused on opinion mining and sentiments analysis," in Programming and Systems (ISPS), 2015 12th International Symposium on.IEEE, vol., no., pp.1-10, 28-30 April 2015.

[2] SADAT, Fatiha, KAZEMI, Farnazeh, et FARZINDAR, Atefeh. Automatic identification of arabic dialects in social media. In : *Proceedings of the first international workshop on Social media retrieval and analysis*. ACM, 2014. p. 35-40..

[3] SAADANE, Houda et HABASH, Nizar. A Conventional Orthography for Algerian Arabic. In : *ANLP Workshop 2015*. 2015. p. 69.

[4] SHOUFAN, Abdulhadi et AL-AMERI, Sumaya. Natural Language Processing for Dialectical Arabic: A Survey. In : *ANLP Workshop 2015*. 2015. p. 36.

[5] HABASH, Nizar et RAMBOW, Owen. MAGEAD: A morphological analyzer and generator for the Arabic dialects. In :Proceedings of the 21st International Conference on Computational Linguistics and the 44th annual meeting of the Association for Computational Linguistics. Association for Computational Linguistics, 2006. p. 681-688.

[6] HABASH, Nizar et RAMBOW, Owen. Morphophonemic and orthographic rules in a multi-dialectal morphological analyzer and generator for arabic verbs. In :*International Symposium on Computer and Arabic Language (ISCAL), Riyadh, Saudi Arabia*. 2007.

[7] SADAT, Fatiha, MALLEK, Fatma, SELLAMI, Rahma, *et al*.Collaboratively Constructed Linguistic Resources for Language Vari-ants and their Exploitation in NLP Applications–the case of Tunisian Arabic and the Social Media. In :*Workshop on Lexical and Grammatical Resources for Language Processing*. 2014. p. 102.

[8] HABASH, Nizar Y., DIAB, Mona T., et RAMBOW, Owen C. Conventional Orthography for Dialectal Arabic (CODA): Principles and Guidelines--Egyptian Arabic-Version 0.7-March 2012. 2014.

[9] GRAFF, David, BUCKWALTER, Tim, JIN, Hubert, *et al.* Lexicon Development for Varieties of Spoken Colloquial Arabic. In :*Proceedings of the Fifth International Conference on Language Resources and Evaluation (LREC)*. 2006. p. 999-1004.

[10] BOUJELBANE, Rahma, BENAYED, Siwar, et BELGUITH, LamiaHadrich. Building bilingual lexicon to create Dialect Tunisian corpora and adapt language model. ACL 2013, 2013, p. 88.

[11] AL-SABBAGH, Rania et GIRJU, Roxana. YADAC: Yet another Dialectal Arabic Corpus. In :*LREC*. 2012. p. 2882-2889.

[12] ELFARDY, Heba et DIAB, Mona T. Sentence Level Dialect Identification in Arabic. In :*ACL (2)*. 2013. p. 456-461.

[13] ALGHAMDI, Mansour, ALHARGAN, Fayez, ALKANHAL, Mohammed, *et al.*Saudi accented Arabic voice bank. *Journal of King Saud University-Computer and Information Sciences*, 2008, vol. 20, p. 45-64.

[14] MOHAMED, Emad, MOHIT, Behrang, et OFLAZER, Kemal. Transforming standard Arabic to colloquial Arabic. In :*Proceedings of the 50th Annual Meeting of the Association for Computational Linguistics: Short Papers-Volume 2*. Association for Computational Linguistics, 2012. p. 176-180.

[15] JEHL, Laura, HIEBER, Felix, et RIEZLER, Stefan. Twitter translation using translation-based cross-lingual retrieval. In :*Proceedings of the Seventh Workshop on Statistical Machine Translation*. Association for Computational Linguistics, 2012. p. 410-421.

[16] ABDUL-MAGEED, Muhammad, DIAB, Mona, et KÜBLER, Sandra. SAMAR: Subjectivity and sentiment analysis for Arabic social media. *Computer Speech & Language*, 2014, vol. 28, no 1, p. 20-37.

[17] HEDAR, Abdel Rahman et DOSS, M. M. MINING SOCIAL NETWORKS'ARABIC SLANG COMMENTS.

[18] SMAÏLI, Kamel, ABBAS, Mourad, MEFTOUH, Karima, *et al.*Building Resources for Algerian Arabic Dialects. In :*15th Annual Conference of the International Communication Association Interspeech*. 2014.

[19] MEFTOUH, Karima, BOUCHEMAL, Nadjette, et SMAÏLI, Kamel. A study of a non-resourced language: an Algerian dialect. In : *SLTU*. 2012. p. 125-132.

[20] BLANC, Olivier et DISTER, Anne. Automates lexicaux avec structure de traits. *Actes de RECITAL*, 2004, p. 23-32.

[21] WOLPER, Pierre. Introduction à la calculabilité. Dunod, 2006.